\documentclass{article}
\pdfoutput=1
\usepackage{arxiv}
\usepackage{lineno,hyperref}
\usepackage{amsmath}
\usepackage{graphicx}
\usepackage{amsfonts}
\usepackage{longtable}
\usepackage{adjustbox}
\usepackage{hhline}
\usepackage{multirow}
\usepackage{algorithm}
\usepackage{algorithmic}
\usepackage{slashbox}
\usepackage[T1]{fontenc}
\usepackage{lscape}
\usepackage{xcolor}
\usepackage{appendix}

\usepackage[utf8]{inputenc} 
\usepackage[T1]{fontenc}    
\usepackage{url}            
\usepackage{booktabs}       
\usepackage{amsfonts}       
\usepackage{nicefrac}       
\usepackage{microtype}      
\usepackage{lipsum}

\title{Adaptive Low-Rank Factorization to regularize shallow and deep neural networks}

\author{Mohammad Mahdi Bejani \\
  Department of Mathematics and Computer Science \\
  Amirkabir University of Technology (Tehran Polytechnic)\\
  Iran\\
  \texttt{mbejani@aut.ac.ir}\\
   \And
  Mehdi Ghatee\footnote{The corresponding author} \\
  Department of Mathematics and Computer Science \\
  Amirkabir University of Technology (Tehran Polytechnic)\\
  Iran\\
  \texttt{ghatee@aut.ac.ir}\\
}

\begin{document}
\maketitle
\begin{abstract}
The overfitting is one of the cursing subjects in the deep learning field. To solve this challenge, many approaches were proposed to regularize the learning models. They add some hyper-parameters to the model to extend the generalization; however, it is a hard task to determine these hyper-parameters and a bad setting diverges the training process. In addition, most of the regularization schemes decrease the learning speed. Recently, Tai et al. \cite{tai2015convolutional} proposed low-rank tensor decomposition as a constrained filter for removing the redundancy in the convolution kernels of CNN. With a different viewpoint, we use Low-Rank matrix Factorization (LRF) to drop out some parameters of the learning model along the training process. However, this scheme similar to \cite{tai2015convolutional} probably decreases the training accuracy when it tries to decrease the number of operations. Instead, we use this regularization scheme adaptively when the complexity of a layer is high. The complexity of any layer can be evaluated by the nonlinear condition numbers of its learning system. The resulted method entitled `'AdaptiveLRF'' neither decreases the training speed nor vanishes the accuracy of the layer. The behavior of AdaptiveLRF is visualized on a noisy dataset. Then, the improvements are presented on some small-size and large-scale datasets. The preference of AdaptiveLRF on famous dropout regularizers on shallow networks is demonstrated. Also, AdaptiveLRF competes with dropout and adaptive dropout on the various deep networks including MobileNet V2, ResNet V2, DenseNet, and Xception. The best results of AdaptiveLRF on SVHN and CIFAR-10 datasets are 98\%, 94.1\% F-measure, and 97.9\%, 94\% accuracy. Finally, we state the usage of the LRF-based loss function to improve the quality of the learning model. 
\end{abstract}

\keywords{Neural Network \and Regularization \and Model Simplification \and Matrix Factorization\and Tensor Factorization.}

\section{Introduction}
In the supervised machine learning, we try to find a learning function $f$ to predict the output of a system by considering its inputs. The complexity of a learning function $f$ can be defined as \cite{dem2015reg}:
\begin{equation}\label{eq:tikhonov-regularization}
R(f) = \int \|\frac{\partial f}{\partial x} \|_2^2dx
\end{equation}
$f$ is complex as much as $R(f)$ is great. When the model complexity is high, a small noise in the input causes a great change in the output and the generalization fails and the overfitting occurs. In the case of deep neural networks, because of their intrinsic complexity, the model tends to memorize the samples and the generalization power reduces \cite{cawley2007preventing}. To solve this problem, different regularization methods are defined to augment a dynamic noise to the model through the training procedure \cite{bejanireview,bejani2019regularized}. One of the most popular techniques is dropout \cite{srivastava2014dropout} and its family \cite{wan2013regularization,kang2018shakeout,khan2018bridgeout,krueger2016zoneout,larsson2016fractalnet,khan2019regularization}. In these methods, in each iteration a subset of weights of the neural network is selected to train. Some of these methods such as \cite{kang2018shakeout,khan2018bridgeout}, impose small changes on the rest of the weights. This family of methods imposes noise on the weights and does not allow the model to memorize the details of the training dataset. \\
However, in many regularization techniques, the noise is imposing blindly to all components of the learning model and they do not pay attention to the time and place of the overfitting \cite{abbasi2016regularized}. This is the reason for the slow convergence in the training of these models. To solve this problem, Abpeikar et al. \cite{abpeikar2020adaptive} proposed an expert node in the neural trees to evaluate the overfitting along with the training, and when it is high, they used regularization. Bejani and Ghatee \cite{bejani2019convolutional}  introduced the adaptive regularization schemes including the adaptive dropout and adaptive weight decay to control overfitting in the deep networks. But their methods did not simplify the structure of the network weights and the learning model became complex in many iterations. However, various matrix decomposition methods such as spectral decomposition, nonnegative matrix factorization, and low-rank factorization have been proposed to summarize the information in the matrix (or tensor)  \cite{symeonidis2016matrix}. For the application of these methods in the data mining fields, one can note to \cite{ng2002spectral}. It seems that, they are also good options for simplifying matrix weights in deep neural networks. In this regard, we find the following attempts: \cite{denton2014exploiting, jaderberg2014speeding, lebedev2014speeding}. In a recent paper, Tai et al. \cite{tai2015convolutional} used low-rank tensor decomposition to remove the redundancy in CNN kernels. Also, Bejani and Ghatee \cite{bejani2020NN} derived a theory to regularize deep networks dynamically by using Singular Value Decomposition (SVD). \\
In continuation of these works, in this paper, we define a new measure based on the condition number of the matrices to evaluate when the overfitting occurs. We also identify which layers of a deep neural network have caused the overfitting problem. To address this problem, we use matrix simplification by decomposing matrices into low-rank matrices. This Low-Rank Factorization that drops out the weights adaptively, is entitled as `'AdaptiveLRF''. This method can compete with the popular dropout and in many cases surpasses dropout. These results will be supported by some experiments on small-size and large-scale datasets, separately. We also visualize the behavior of AdaptiveLRF on a noisy dataset. Then, on dataset CIFAR-100, we show the performance of AdaptiveLRF using VGG-19. Finally, the results of AdaptiveLRF are compared with some famous regularization schemes including dropout methods  \cite{kang2018shakeout,khan2018bridgeout,krueger2016zoneout,larsson2016fractalnet,khan2019regularization}, adaptive dropout method \cite{bejani2019convolutional}, weight decay with and without augmentation methods \cite{engstrom2017rotation,kwasigroch2017deep,wkasowicz2017computed,galdran2017data}. \\
In what follows, we present some preliminaries in Section 2. In Section 3, the AdaptiveLRF is expressed. In Section 4 we present the empirical studies. The final section ends the paper with a brief conclusion.

\section{Preliminaries}
The overfitting of a supervised learning model such as a neural network is related to the condition number of the following nonlinear system:
\begin{equation}\label{eq:learning_label}
\sum_{i=1}^T\|f(x_i,\{w_1,...,w_L\}) - y_i\|_F^2 = 0,
\end{equation}
where $f$ is the output of the neural network with $L$ layers. $w_l$ is the weight matrix (or tensor) of the layer $l.$ $T$ is the number of training samples, and $(x_i,y_i)$ is the pair of the inputs and outputs of the $i^{th}$ sample. After solving this nonlinear system and finding $w_i$s, the learning model can be used to predict the output for any unseen data. In numerical algebra, it was shown that the condition number of a system is dependent directly on the stability of the solution \cite{datta2010numerical}. Really, when the condition number is great (very greater than 1) the sensitivity of the system over the noise is very high and so the generalization ability of the learning model decreases significantly. Thus, it is a good idea to evaluate the complexity of the learning model by condition number \cite{bejani2020NN}. The condition number can be defined for linear and nonlinear systems. For a linear system $Ax=b,$ where $A\in \mathbb{R}^{m\times n}$, $x\in \mathbb{R}^{n}$, and $b\in \mathbb{R}^{m}$, the condition number is defined as $\kappa(A) =\|A\|\|A^{-1}\|$ \cite{datta2010numerical}. Really, when the condition number is great (very greater than 1) the sensitivity of the system over the noise. Also, for the non-linear system $f(x) = y$ where $f$ is a non-linear vectorized function, one can use the following formula \cite{trefethen1997numerical}:
\begin{equation}\label{eq:non-linear-condition-number}
\kappa(f(\theta)) = \frac{\|J(\theta)\|_F \|\theta\|_F}{\|f(\theta)\|_F},
\end{equation}
where $\|.\|_F$  is Frobenius norm, $\theta$ is parameters of $f$, and $J(\theta)$ is Jacobian matrix of $f$ respect to $\theta$.

\subsection{Matrix factorization}
In this part, we discuss popular matrix factorization (decomposition) and show their ability to improve the system stability.  Consider an arbitrary matrix $A$ that is factorized into $r$ matrices $B_i$ and $A = \prod_{i=1}^r B_i$. In some instances, LU decomposition, Cholesky decomposition, Singular Value Decomposition (SVD), nonnegative matrix decomposition, binary decomposition, can be used to determine the factors \cite{elden2019matrix}. Now, we focus on the low-rank factorization that approximates any matrix $A$ with two lower rank matrices $W$ and $H$. To improve the approximation, the following optimization problem can be solved:
\begin{equation}\label{eq:LRF}
\min_{W,H} \|A - WH\|_F
\end{equation}
When $W$ and $H$ are two vectors, their ranks are 1 and the matrix $A$ is factorized to two matrices with the lowest ranks. We refer to this factorization with LRF. Thus, when $A\in R^{n,m},$ LRF factorize it into two matrices $W\in R^{n,1}$ and $H\in R^{1,m}.$ To satisfy \ref{eq:LRF}, we should solve a nonlinear system with $m + n$ variables and $m.n$ equation. See \cite{nimfa2020fact} for an implementation. 

\subsection{Tensor factorization}
There are two main approaches to factorize a tensor; explicit and implicit. In the explicit factorization of any tensor $T$, we try to find $r$ sets of vectors $a_i$, $b_i$ and $c_i$ such that $\sum_{i=1}^r (a_i b_i^T) \odot c_i$ approximates $T$, where $\odot$ is the tensor production. By minimizing $\|T - \sum_{i=1}^r (a_i b_i^T) \odot c_i\|_2^2,$ we can factorize $T$ into $r$ components \cite{tai2015convolutional}. However, the explicit tensor decomposition is an NP-hard problem\cite{hillar2013most}. Therefore, this type of decomposition is not the best way for the regularization of deep networks. Instead, in implicit factorization, we try to apply the matrix factorization methods directly. To this aim, any tensor $T$ is sliced into some matrices and on every matrix, we apply the matrix factorization. The results show the efficiency of this approach for deep learning regularization.
\subsection{Visualization of factorization effects}
\begin{figure}
\centering
\includegraphics[scale=0.45]{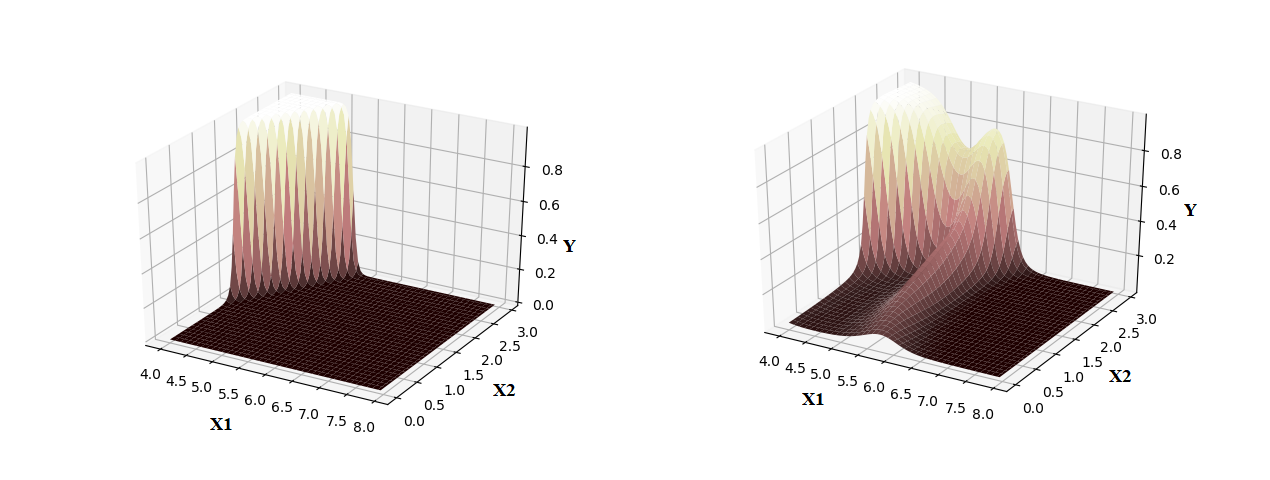}
\caption{The learning surface that trained by a perceptron neural network (MLP) on Iris dataset without noise (left figure) and on noisy Iris dataset (right figure).}
\label{fig:original-network-function}
\end{figure}

\begin{figure}
\centering
\includegraphics[scale=0.45]{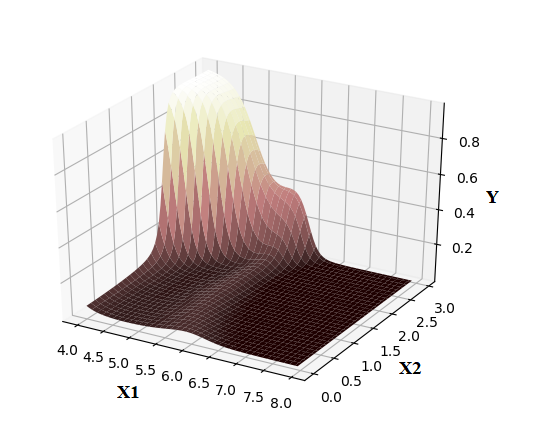}
\caption{The regularized surface made by a perceptron neural network (MLP) on noisy Iris dataset when LRF is used on weighting matrices.}
\label{fig:factorization-network-function}
\end{figure}

To visualize the effect of matrix and tensor factorization as the regularization method, we designed a test to show how they can improve the learning functions. To this end, we used an artificial noisy dataset based on Iris dataset \cite{anderson1936species} and constructed a surface to learn these noisy data by a perceptron neural network with 3 hidden layers. Fig.\ref{fig:original-network-function} shows two surfaces that are trained by the original and noisy datasets separately. As one can see, the learning model is over-fitted because of noisy data. Now, we use LRF on the weighting matrices of the corresponding neural network to regularize this learning model.  Fig.\ref{fig:factorization-network-function} shows the new surface. It is trivial that the regularized network is more similar to the original learning model that has not destroyed by noisy data. Also, the model is simpler.

\section{AdaptiveLRF details}
AdaptiveLRF is a regularization technique that is developed for deep neural networks but not limited to these networks. The fundamental steps of this technique are:
\begin{enumerate}
\item Detecting the overfitting in continuous steps,
\item Identifying the matrices with great effect on overfitting,
\item Selecting some over-fitted matrices randomly,
\item Using LRF to regularize the over-fitted matrices and the tensors.
\end{enumerate}
To present the details, we need to consider several important points. Noting that, a powerful method for regularization needs to evaluate the overfitting dynamically \cite{bejani2019convolutional}. When the overfitting is small, the learning procedure can be continued, else, the overfitting should be solved by a regularization method. Such a scheme saves the training speed and increases the generalization ability. The dynamic overfitting can be evaluated b the following criterion:
\begin{equation}\label{eq:over-fitting-strict-computation}
v(t) = \frac{Error_{Validation}(t)}{Error_{Train}(t)},
\end{equation}
where $t$ is the iteration number. It is worthwhile to note that $v(t)$ has an oscillatory behavior and iteratively decreases and increases. Therefore, the average of the last $p = 3$ $v(t)$ can be considered. $p$ is named as \textit{patient of regularization}. When the overfitting is recognized, the cause of the overfitting must be identified and treated. Because of the layered architecture of deep networks, it is possible to find some of the layers that cause overfitting. The weights of these layers should be regularized to miss some details of data captured by the weighting matrices. However, the major trend of data should be prevented. This leads to a softer surface (as mentioned in Fig.\ref{fig:factorization-network-function}). 

At first glance, finding a sub-set of the layers with the highest effect on overfitting is hard. Instead, we return to the training system \label{eq:learning_label} and compute the complexity of each layer by its condition number \ref{eq:non-linear-condition-number}. Denote the  condition number of $l^{th}$ layer with $\kappa(f_l)$. We are ready to regularize the weighting matrices with great $\kappa(f_l).$ But, the experiments show that regularization on every over-fitted matrix increases the processing time. Thus, similar to dropout \cite{srivastava2014dropout}, we define a random test by using $Bernoulli(\Gamma(w_i))$ distribution. When the produced random parameter is less than the following normalized parameter, we use LRF regularization to simplify the weighting matrices:
\begin{equation}\label{eq:normalize_the_cd}
\Gamma(f_l) = \frac{\kappa(f_l)}{\max_i \kappa(f_i)},
\end{equation}
To follow the LRF regularization, for the layers with the dense weights and the last convolution layer, the weighting matrices can be approximated by LRF. For the convolution layers with tensor structure, they are sliced to some small matrices with the size of the filter and the number of filters. Then, the LRF approximations are defined on these matrices.

The topic is discussed in the empirical results section. The summarization of the training algorithm with AdaptiveLRF is stated in Algorithm 1.

\begin{algorithm}
\caption{Training Algorithm with AdaptiveLRF}
\begin{algorithmic}[1]
\REQUIRE $\alpha_t$: Step-size
\REQUIRE $D_t$: Improvement Direction
\REQUIRE $\theta_t \leftarrow \{W_t,b_t\}$: Weights and biases of the neural network.
\REQUIRE $E(\theta_t)$: Error function of the network.
\REQUIRE $W$: number of the weights in the network.
\REQUIRE $L$: number of the trainable layers in the network.

\STATE $t \leftarrow 0$
\WHILE{$\theta_t$ does not converge}
	\STATE $g_{t}\leftarrow \bigtriangledown_\theta E(\theta_t)$
	\STATE $D_{t} \leftarrow DescentDirection(g_{t}, \alpha_t)$ that is a function to return an improvement direction based on the inputed gradient and $\alpha_t$.
	\STATE $\theta_{t+1} \leftarrow \theta_t -D_{t+1}$ 
	\STATE $E_{Train}(t)$ is evaluated as the error over the training samples ($Data_{Train}$).
	\STATE $E_{Validation}(t)$ is evaluated as the error over the validation samples ($Data_{Validation}$).
	\STATE $v(t) \leftarrow \dfrac{E_{Train}(t)}{E_{Validation}(t)}.$ 
	\IF{$v(t)$ is high}
		\FOR{ \text{all layers} $l \in \{1,...,L \}$ and any weights tensor $W_{l}$ and bias vector $b^l$} 
			\STATE Compute the $\kappa(f_l)$ by Eq. \ref{eq:non-linear-condition-number}.
		\ENDFOR
		\FOR{ \text{all layers} $l \in \{1,...,L \}$ and any weights tensor $W_{l}$} 
			\STATE Compute $\Gamma(f_l)$ based on Eq. \ref{eq:normalize_the_cd}
			\STATE $r \leftarrow rand(0,1)$
			\IF{$r \leq \Gamma(f_l)$}
				\STATE $W_n \leftarrow$ approximation of $W_n$.
			\ENDIF
		\ENDFOR
		\STATE $b^{(l)}_{t+1} \leftarrow b^{(l)}_{t}$
	\ENDIF
	\STATE $t \leftarrow t + 1$
\ENDWHILE
\RETURN{$\theta_{t}$}
\end{algorithmic}
\end{algorithm}

\section{Empirical studies}
In this section, the numerical results are shown on AdaptiveLRF and compared them with other regularization methods. Also, we check its power to control the overfitting in the different datasets by using shallow and deep networks. The implementation of AdaptiveLRF can be found \href{https://github.com/mmbejani/AdaptiveLRF}{here}\footnote{https://github.com/mmbejani/AdaptiveLRF}.

\subsection{Effect of condition number in AdaptiveLRF}
To present the effect of condition number expressed in Eq. \ref{eq:normalize_the_cd} in the performance of AdaptiveLRF,  consider the following scenarios:

\begin{enumerate}
\item The $k$ first layers of the network are used for regularization when the overfitting occurs. In this scenario, the weight matrices of the first $d$ layers are factorized by LRF and their approximations are substituted as the new weights matrices.
\item The last $d$ layers of the network are used for regularization and so on.

\item This scenario is the combination of the random selection and standard AdaptiveLRF that are presented in Algorithm 1.
\end{enumerate}

To compare these scenarios, VGG-19 network was trained on CIFAR-100. To trace these scenarios, we define the following criterion namely summation of normalized condition number (SNCN):

\begin{equation}
SNCN(f) = \sum_{l=1}^L \Gamma(f_l),
\end{equation}

where $\Gamma(f_l)$ is defined in Eq.\ref{eq:normalize_the_cd} as the condition number of layer $l$. The smaller this criterion in different iterations, the greater the network's stability against overfitting.

\begin{figure}
\centering
\includegraphics[scale=0.42]{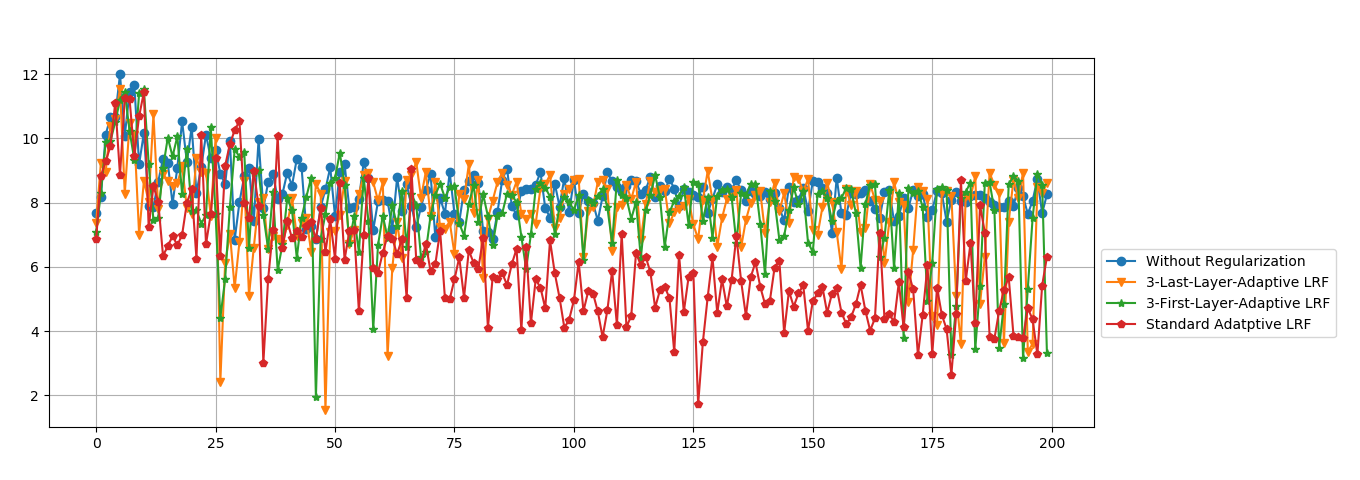}
\caption{SNCN criterion of VGG-19 on CIFAR-100 in each epoch for three scenarios. (For interpretation of the references to color in this figure legend, the reader is referred to the web version of this article.)}
\label{fig:scnarios_vgg_cd}
\end{figure}

\begin{figure}
\centering
\includegraphics[scale=0.42]{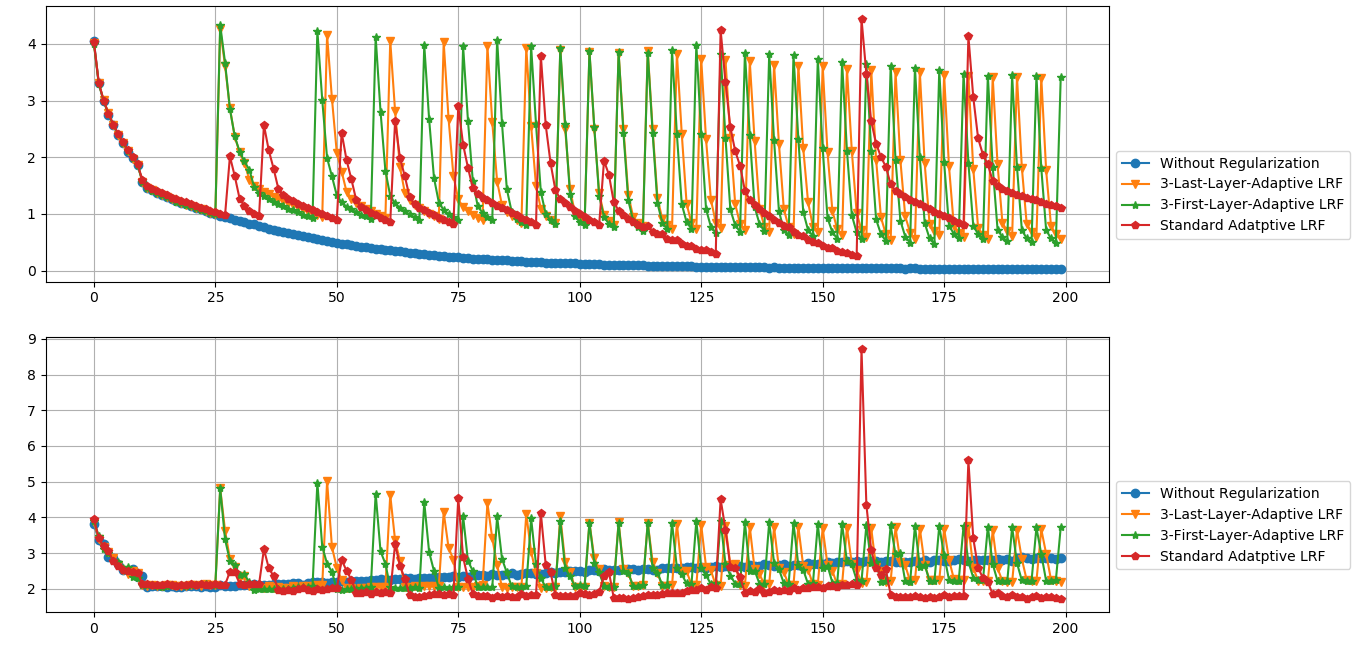}
\caption{The training and testing loss values of VGG-19 on CIFAR-100 in each epoch for three scenarios. (For interpretation of the references to color in this figure legend, the reader is referred to the web version of this article.)}
\label{fig:scnarios_vgg_per}
\end{figure}

In figures \ref{fig:scnarios_vgg_cd} and \ref{fig:scnarios_vgg_per}, the performance of the VGG-19 for the presented scenarios are presented. As one can see, the performance of AdaptiveLRF for the third scenario is better compared with the others in terms of SNCN, training loss, and testing loss values. 

\begin{figure}
\includegraphics[width=0.5\textwidth]{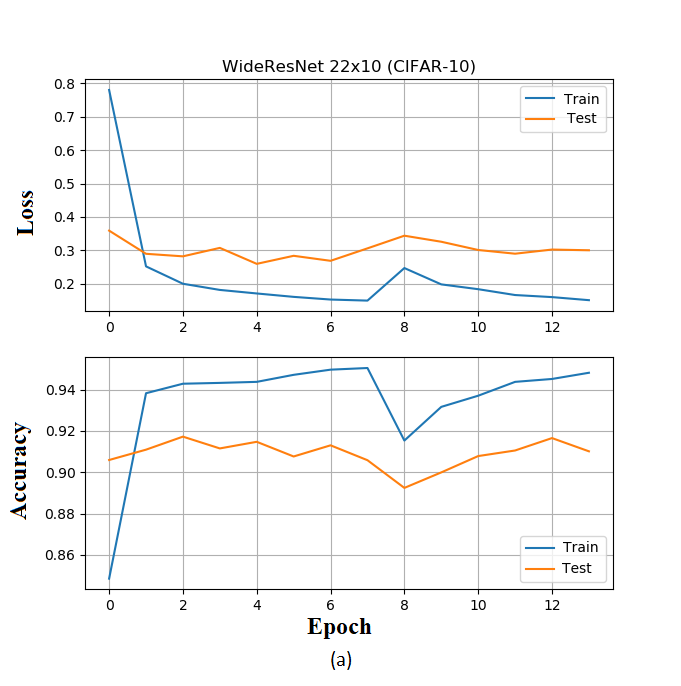}
\includegraphics[width=0.5\textwidth]{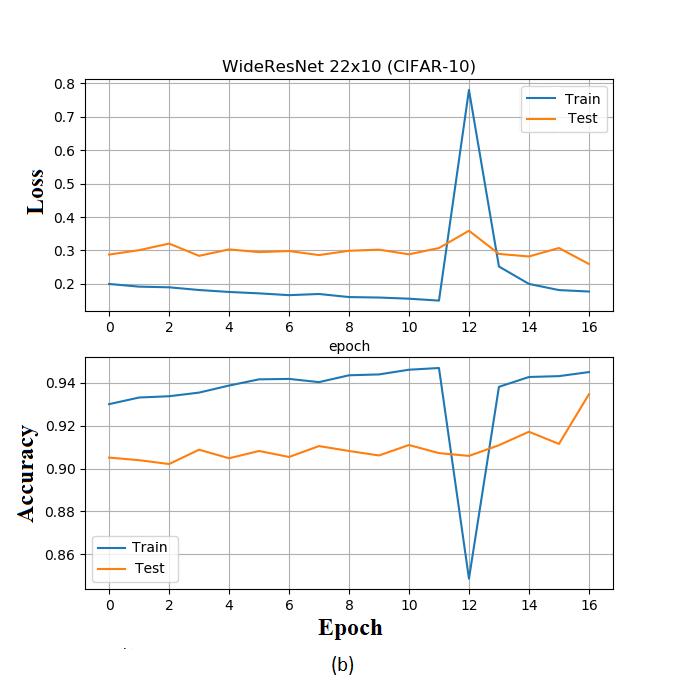}
\caption{The regularization effect of AdaptiveLRF on over-fitted epochs. (For interpretation of the references to color in this figure legend, the reader is referred to the web version of this article.)}
\label{fig:wrn-overfitting-cifar10-nmf-1}
\end{figure}

In addition, one can see the leap of the loss function in training and testing results is different. Two snapshots of them are shown in Fig. \ref{fig:wrn-overfitting-cifar10-nmf-1}. These snapshots are extracted from the learning procedure of Wide-Resnet on CIFAR-10. As one can see, the AdaptiveLRF has affected by the parts of the network, where the model is over-fitted. Thus, AdaptiveLRF simplifies the model when the model is over-fitted and has a low affect on the other parts. This means that the useful information is seldom eliminated.

\subsection{Performance of AdaptiveLRF on Shallow Networks}
In this part, we show the results of AdaptiveLRF on the shallow networks applying different datasets. The used shallow networks have at most 5 layers and the layers are composed of dense layer and one dimension convolution layer. The results of AdaptiveLRF for this network are compared with the case that no regularization is used. Also the results of dropout with $p=0.1,0.2,0.3$ are presented. Each experiment is repeated 5 times and their average is presented. The results are shown in Table \ref{tab:small_dataset_comp}. As one can see, in datasets Arcene, BCWD, and IMDB Reviews, AdaptiveLRF can defeat others. In addition, in datasets BCWD whose levels of overfitting are low, AdaptiveLRF is somewhat weaker than the case of non-regularization. Really, the $v(t)$ for these datasets belong to $[0.9,1.9]$.

\begin{table}
\begin{center}
\caption{The comparison of the performance of the different regularization methods on a shallow network.}
\label{tab:small_dataset_comp}
\begin{tabular}{c|c|c|c|c|c}
\hhline{======}
Dataset Name & Regularization & Train A & Train L & Test A & Test L \\ 
\hhline{======}
\multirow{5}{*}{Arcene} & None & 99.4\%& $6.04\times 10^{-3}$ & 74.6\%& 0.202\\ \cline{2-6}
& Dropout (0.1) & 60.6\% & $3.87\times 10^{-1}$ & 59.2\% & 0.406\\ \cline{2-6}
& Dropout (0.2) & 60.0\% & $3.80\times 10^{-1}$ & 59.0\% & 0.411\\ \cline{2-6}
& Dropout (0.3) & 61.1\% & $3.72\times 10^{-1}$ & 58.0\% & 0.413\\ \cline{2-6}
& AdaptiveLRF & 97.4\% & $6.80\times 10^{-2}$ & \textbf{85.8\%} & \textbf{0.125}\\ \hline
\multirow{5}{*}{BCWD} & None & 98.7\% & $1.03\times 10^{-2}$ & 95.0\% & 0.038 \\ \cline{2-6}
& Dropout (0.1) & 99.8\% & $3.02\times 10^{-3}$ & 95.3\% & 0.037\\ \cline{2-6}
& Dropout (0.2) & 93.1\% & $4.50\times 10^{-2}$ & 90.4\% & 0.068\\ \cline{2-6}
& Dropout (0.3) & 90.8\% & $6.60\times 10^{-2}$ & 86.5\% & 0.112\\ \cline{2-6}
& AdaptiveLRF & 97.6\% & $1.86\times 10^{-2}$ & \textbf{95.6\%} & \textbf{0.032}\\ \hline
\multirow{5}{*}{BCWP} & None & 98.3\% & $1.22\times 10^{-2}$ & \textbf{76.4\%} & \textbf{0.205}\\ \cline{2-6}
& Dropout (0.1) & 97.1\% & $2.73\times 10^{-2}$ & 75.0\% & 0.229\\ \cline{2-6}
& Dropout (0.2) & 95.0\% & $3.52\times 10^{-2}$ & 71.9\% & 0.220\\ \cline{2-6}
& Dropout (0.3) & 88.0\% & $8.96\times 10^{-2}$ & 73.2\% & 0.211\\ \cline{2-6}
& AdaptiveLRF & 89.2\% & $8.63\times 10^{-2}$ & 74.6\% & 0.210\\ \hline

\multirow{5}{*}{IMDB Reviews} & None & 100.0\% & $7.60\times 10^{-4}$ & 85.3\% & 0.396\\ \cline{2-6}
& Dropout (0.1) & 99.7\% & $1.14\times 10^{-2}$ & 84.5\% & 0.542\\ \cline{2-6}
& Dropout (0.2) & 99.7\% & $1.14\times 10^{-2}$ & 84.5\% & 0.542\\ \cline{2-6}
& Dropout (0.3) & 99.7\% & $1.14\times 10^{-2}$ & 84.5\% & 0.542\\ \cline{2-6}
& AdaptiveLRF & 99.4\% & $3.54\times 10^{-2}$ & \textbf{85.6\%} & \textbf{0.354}\\ 
\hhline{======}
\end{tabular}
\end{center}
\end{table}

\subsection{Performance of AdaptiveLRF on deep networks}
In this part, AdaptiveLRF is evaluated on the different popular standard datasets and CNNs. The performance of AdaptiveLRF is illustrated with augmentation and without augmentation.

\subsubsection{Comparison Performance}
We compare this method with the other regularization methods including weight decay\cite{krogh1992simple}, dropout \cite{srivastava2014dropout} and adaptive weight decay and adaptive dropout \cite{bejani2019convolutional}. We use popular networks configuration such as MobileNet V2 \cite{sandler2018mobilenetv2}, ResNet V2 \cite{he2016identity}, DenseNet \cite{huang2017densely} and Xception \cite{chollet2017xception}. Also, we augment input images with Cutout method\cite{devries2017improved}. In all of the experiments, the Adam optimization algorithm are used and the number of maximum epochs is fixed for each dataset, individually. We also use SVHN \cite{netzer2011reading} and CIFAR-10\cite{krizhevsky2009learning} as the datasets. In what follows, we present the results.

\subsubsection{SVHN}
SVHN is an image dataset containing about 600,000 images for the training and 26,000 images for the testing. We consider 200 epochs and evaluate the performance of the different networks with augmentation and without augmentation in Table \ref{tab:svhn-results}. The augmentation consists of the following operations: 

\begin{itemize}
\item Rotation between $-20^{\circ}$ to $20^{\circ}$.
\item Transition the pixels between $-6$ to $6$.
\item Using Cutout with probability $0.5$\cite{devries2017improved}.
\end{itemize}
Because of the high number of training samples, the probability of overfitting of the deep models on SVHN dataset is low (The small performance difference between with augmentation and without augmentation shows that). Therefore, as one can see in Table \ref{tab:svhn-results} the results when regularization is used and without regularization, is close. Besides, sometimes using a regularization scheme causes that the performance decreases (MobileNet with weight decay). However, AdaptiveLRF can bet all of the regularization schemes because it acts when the overfitting appears, and in this dataset that the overfitting level is so low, AdaptiveLRF affects the model lower than the others, therefore, AdaptiveLRF can reach to the better performance.

\begin{table}
\begin{center}
\caption{The performance of the different deep networks on SVHN with the different regularization scheme (Bold values show the best accuracies)}
\label{tab:svhn-results}
\resizebox{\textwidth}{!}{\begin{tabular}{c|c|c|c|c|c|c|c|c|c|c|c|c}
\hhline{=============}
\multicolumn{13}{c}{Without Augmentation} \\ \hline
\multirow{2}{*}{\backslashbox{Model}{Regularization}} & \multicolumn{2}{c|}{AdaptiveLRF} & \multicolumn{2}{c|}{Weight Decay} & \multicolumn{2}{c|}{Dropout} & \multicolumn{2}{c|}{Adaptive WD}&\multicolumn{2}{c|}{Adaptive Dropout}&\multicolumn{2}{c}{None}\\ 
\cline{2-13}
& A & F & A & F & A & F & A & F& A & F& A & F
\\ \hhline{=============}
MobileNet V2 & \textbf{96.8} & \textbf{97.0} & 93.3 & 93.4 & 93.0 & 93.1 & 95.8 & 96.0 & 94.7 & 94.9 & 96.6 & 96.6\\ \hline
ResNet V2 & \textbf{96.6} & \textbf{96.7} & 95.8 & 95.9 & 96.1 & 96.1 & 95.8 & 95.8 & 96.1 & 96.2 & 96.4 & 96.5\\ \hline
DenseNet & \textbf{97.8} & \textbf{97.9} & 96.5 & 96.6 & 97.1 & 97.1 & 97.0 & 97.2 & 97.3 & 97.4 & 96.4 & 96.4\\ \hline 
Xception & \textbf{97.9} & \textbf{98.0} & 96.5 & 96.6 & 97.1 & 97.2 & 97.3 & 97.3 & 97.2 & 97.3 & 97.2 & 97.4\\ \hline
\multicolumn{13}{c}{With Augmentation} \\ \hline
MobileNet V2 & 96.8 & 97.0 & 95.4 & 95.5 & 89.1 & 89.4 & 95.6 & 95.7 & 95.2 & 95.3 & \textbf{97.2} & \textbf{97.2}\\ \hline
ResNet V2 & \textbf{97.4} & \textbf{97.4} & 97.1 & 97.1 & 97.1 & 97.1 & 95.7 & 95.7 & 96.1 & 96.2 & 97.2 & 97.3\\ \hline
DenseNet & \textbf{97.9} & \textbf{98.0} & 96.9 & 97.0 & 95.1 & 95.3 & 97.2 & 97.4 & 95.6 & 95.7 & 97.6 & 97.7\\ \hline 
Xception & \textbf{97.9} & \textbf{98.0} & 97.4 & 97.4 & 97.2 & 97.4 & 97.4 & 97.5 & 97.6 & 97.7 & 97.6 & 97.7
\\ \hhline{=============}
\end{tabular}}
\end{center}
\centering
{\tiny * The A and F are accuracy, F-measure.}
\end{table}

\subsubsection{CIFAR-10}
The CIFAR-10 \cite{krizhevsky2009learning} is smaller than SVHN with the same number of classes. We evaluate and compare AdaptiveLRF with other regularization schemes on this dataset with augmentation and without augmentation. The augmentation strategies for CIFAR-10 is as following:
\begin{itemize}
\item Rotation between $-20^{\circ}$ to $20^{\circ}$.
\item Transition the pixels between $-3$ to $3$.
\item Horizontal flip the images by probability 0.5.
\item Using Cutout with probability $0.5$\cite{devries2017improved}.
\end{itemize}
We illustrate the results in Table \ref{tab:cifar10-results}. The reported results are achieved after 200 epochs.
 As one can see, AdaptiveLRF can overcome the other regularization schemes in most of the cases. Besides, the difference between accuracies when using augmentation is lower than when using raw data. This shows that the level of overfitting is decreased when the data is augmented. However, by decreasing the level of the overfitting the effect of AdaptiveLRF decreases and reaches better performance respect to others.

\begin{table}
\begin{center}
\caption{The performance of the different deep networks on CIFAR-10 with the different regularization scheme (Bold values show the best accuracies)}
\label{tab:cifar10-results}
\resizebox{\textwidth}{!}{\begin{tabular}{c|c|c|c|c|c|c|c|c|c|c|c|c}
\hhline{=============}
\multicolumn{13}{c}{Without Augmentation} \\ \hline
\multirow{2}{*}{\backslashbox{Model}{Regularization}} & \multicolumn{2}{c|}{AdaptiveLRF} & \multicolumn{2}{c|}{Weight Decay} & \multicolumn{2}{c|}{Dropout} & \multicolumn{2}{c|}{Adaptive WD}&\multicolumn{2}{c|}{Adaptive Dropout}&\multicolumn{2}{c}{None}\\ 
\cline{2-13}
& A & F & A & F & A & F & A & F& A & F& A & F
\\ \hhline{=============}
MobileNet V2 & \textbf{75.0} & \textbf{75.2} & 71.3 & 71.5 & 74.5 & 74.6 & 71.7 & 71.8 & 74.8 & 74.8 & 70.9 & 71.1\\ \hline
ResNet V2 & 73.9 & 74.1 & 73.5 & 73.7 & 74.6 & 74.6 & 74.6 & 74.6 & \textbf{74.7} & \textbf{74.9} & 72.7 & 72.9\\ \hline
DenseNet & 75.1 & 75.1 & 74.1 & 74.1 & 75.3 & 75.8 & 74.6 & 74.8 & \textbf{75.3} & \textbf{75.5} & 73.8 & 73.9\\ \hline 
Xception & \textbf{75.7} & \textbf{75.8} & 73.5 & 73.6 & 74.0 & 74.1 & 74.5 & 74.5 & 74.5 & 74.7 & 71.8 & 71.0\\ \hline
\multicolumn{13}{c}{With Augmentation} \\ \hline
MobileNet V2 & \textbf{91.9} & \textbf{92.3} & 91.6 & 91.6 & 91.7 & 91.8 & 91.8 & 91.9 & 91.8 & 92.0 & 91.1 & 91.4\\ \hline
ResNet V2 & \textbf{92.5} & 92.5 & 92.3 & 92.6 & \textbf{92.5} & 92.6 & 92.4 & 92.5 & \textbf{92.5} & \textbf{92.7} & 92.1 & 92.3\\ \hline
DenseNet & \textbf{94.0} & \textbf{94.1} & 93.0 & 93.2 & 93.5 & 93.5 & 93.2 & 93.2 & 93.8 & 94.0 & 93.0 & 93.0 \\ \hline 
Xception & 91.5 & 91.7 & 90.7 & 90.8 & \textbf{91.8} & \textbf{92.0} & 91.1 & 91.2 & 91.7 & \textbf{92.0} & 91.7 & \textbf{92.0}
\\ \hhline{=============}
\end{tabular}}
\end{center}
\centering
{\tiny * The A, F are accuracy and F-measure.}
\end{table}

\section{An improvement on AdaptiveLRF by the aid of LRF-based loss function}
The results showed that AdaptiveLRF prefers on the other regularizers for shallow networks and can compete with other adaptive dropout variations. However, the results of AdaptiveLRF in the deep networks were not the best when the model complexity is high. Recently, Bejani and Ghatee \cite{bejani2020NN} proved a new theory for adaptive SVD regularization (ASR). They used the following loss function to accelerate the convergence of the learning problem:
\begin{equation}
E^*(\theta) = E(\theta) + \gamma \|\theta - \theta^*_{Estimated}\|_F^2
\end{equation}
where $\theta=\{W, b\}$ denotes the synaptic weights and the bias vector of a neural network and $\theta^*_{Estimated}=\{W^*_{Estimated}, b^*_{Estimated}\}$ is estimated by the best synaptic weights on the validation dataset. $\gamma$ is used for regularization. They minimized this loss function by using their  SVD approximation of $\theta^*$ . Instead, we can use LRF to approximate this term. Thus, we can use 
the following `'LRF-based loss function'' in our training model:
\begin{equation}
E^*(\theta) = E(\theta) + \|\theta - LRF(\theta^*)\|_F^2
\end{equation}
Based on the initial results, this modification can improve the quality of learning for different deep neural networks. We will present the details of this experiments soon.
\section{Conclusion and future directions}
In this paper, we discussed the effects of an adaptive low-rank factorization for neural network regularization entitled AdaptiveLRF. This regularization scheme was not implemented for all layers, which is different from \cite{tai2015convolutional}. Instead, the conditional number of the synaptic weights for each layer was evaluated and when it was high, the low-rank approximation of the matrices was substituted. This idea was used to retrieve the information of synaptic weights. The proposed AdaptiveLRF can find a stable solution for the learning problem. We showed the results of this scheme on two categories of shallow and deep neural networks. The results showed that AdaptiveLRF prefers on the other regularizers for shallow networks and can compete with other adaptive dropout variations. However, the results of AdaptiveLRF in the deep networks can be improved by using an adaptive plan similar to ASR \cite{bejani2020NN}. We will present the results of AdaptiveLRF together adaptive LRF-based loss function in future work. Also, overfitting is very important in many machine learning branches and it is necessary to solve them by using context knowledge. The effects of AdaptiveLRF for shallow and deep neural networks in these branches should be evaluated. For future works, one can focus on AdaptiveLRF for solving overfitting in neural networks that are used for feature extraction \cite{pashaei2019convolution}, sensors fusion \cite{bejani2018context,eftekhari2018hybrid}, data visualization \cite{bejani2019convolutional}, and ensemble learning \cite{abpeykar2019ensemble}. Since, in \cite{bejani2019regularized} the importance of overfitting in transportation problems has been highlighted, we encourage the researchers to implement AdaptiveLRF in the transportation problems \cite{ghatee2019smartphone}.

\bibliography{AdaptiveLRF}

\appendix

\end{document}